%% file: 00_main.tex
\pdfoutput=1

\documentclass[11pt]{article}

\usepackage{ACL2023}

\usepackage{times}
\usepackage{latexsym}

\usepackage[T1]{fontenc}

\usepackage[utf8]{inputenc}

\usepackage{microtype}

\usepackage{array}
\newcolumntype{H}{>{\setbox0=\hbox\bgroup}c<{\egroup}@{}}

\usepackage{inconsolata}

\usepackage{dirtytalk}
\usepackage{graphicx}
\usepackage{booktabs}
\usepackage{multirow}

\newcommand{\todot}[1]{}
\newcommand{\uvp}[1]{}

\title{Tokenizing Crosslingual Homographs}

\author{Rotem Brillant~~~~~~~Yuval Pinter \\
  Stein Faculty of Computer and Information Science \\
  Ben-Gurion University of the Negev \\
  Beer Sheva, Israel \\
  \texttt{\{brillant@post,uvp@cs\}.bgu.ac.il} \\}

\begin{document}
\maketitle
\begin{abstract}
Multilingual language models rely on shared subword vocabularies to represent multiple languages within a limited number of token units.
While such sharing is often useful, it can also create cases in which identical surface forms are treated too uniformly across languages, even when their meanings or usage differ.
We investigate this limitation through cross-lingual homographs and false friends, and examine whether introducing language information earlier in the tokenization process can improve their treatment.
We propose a simple tokenizer-level intervention based on language cues: language-specific characters replacing initial characters of shared-vocabulary words, reducing common identity during vocabulary construction.
In intrinsic analysis, we find through tokenizer-level statistics that BPE and UnigramLM often treat cross-lingual homographs in a largely language-agnostic way, whereas the context-sensitive SaGe tokenizer diverges more strongly; our intervention removes this gap.
In downstream English-to-X machine translation, our cues yield modest improvements in several settings, especially under BPE, although the effect is not consistent across all languages and evaluation sets.
Overall, the findings suggest that adding lightweight language information at the tokenizer level is a promising direction for further exploration.
\end{abstract}

\section{Introduction}
\label{sec:introduction}
\input{01_intro}

\section{Homographs in Tokenization}
\label{sec:analysis}

\input{02_analysis}

\section{Language Cues for Homographs}
\label{sec:approach}
\input{03_approach}

\section{Evaluation}
\label{sec:exp}
\input{04_experiments}

\section{Conclusion}
\label{sec:conc}
\input{06_conclusion}


\section*{Limitations}

One limitation of this work is its experimental scope. The downstream evaluation is restricted to English-to-L2 machine translation task for a specific set of languages, most of which are fairly related and use Latin-based scripts. This means that the findings should be interpreted with caution, since it is not yet clear how well the same behavior would carry over to more distant language pairs, other writing systems, or broader multilingual settings.

The proposed cueing method is also intentionally narrow: it applies only to a selected set of cross-lingual homographs, and therefore does not address other forms of potentially problematic token sharing. Other forms of cross-lingual ambiguity may not be captured by this design. At the same time, the empirical gains are generally modest and not fully consistent across tokenizers, datasets, and language pairs. Taken together, these limitations suggest that the current results are best viewed as encouraging initial evidence rather than as a complete solution. Future work will be needed to test whether similar ideas remain useful in larger multilingual models, on other tasks, and in more general language-aware tokenization schemes.

\section*{Acknowledgements}
This research was supported by the Israel Science Foundation (grant No. 1166/23).

\bibliography{anthology-1,custom}
\bibliographystyle{acl_natbib}

\newpage

\appendix

\input{99_appendix}

\end{document}

%% file: 01_intro.tex
Multilingual large language models have significantly advanced natural language processing by enabling a single model to represent multiple languages within a shared parameter space~\citep{aharoni-etal-2019-massively, conneau2020unsupervisedcrosslingualrepresentationlearning}.
A central component of these models is the tokenization process, which constructs a subword vocabulary used to encode textual input across languages.
Tokenization plays a critical role in balancing computational efficiency with linguistic expressiveness, particularly in multilingual settings~\citep{limisiewicz-etal-2023-tokenization}.

\input{fig_cues_intro}

Because vocabulary size is limited, selecting an appropriate tokenizer configuration becomes a key architectural decision~\citep{liang-etal-2023-xlm}.
Multilingual contexts introduce unique design considerations, such as how vocabulary capacity should be distributed across languages, whether to train a shared tokenizer or aggregate language-specific ones, and how increasing the number of languages affects cross-lingual transfer performance.

In a shared vocabulary, an important challenge emerges when training tokenizers on multiple related languages: the presence of shared tokens.
Related languages often contain identical surface forms that are assigned a single subword unit in the learned vocabulary.
An example is the word \textit{barn}, shared between English and Swedish.
These shared tokens receive a single embedding representation that is jointly trained across languages during pre-training~\citep{vernikos-popescu-belis-2021-subword-mapping}.
However, this shared representation may not equally capture the meanings and usage patterns of all contributing languages.
Such imbalance can arise due to differences in corpus size, token frequency, or contextual usage patterns across languages.

We investigate the shared-token phenomenon through a specific class of words: cross-lingual homographs, words that share identical surface forms across different languages.
While many such words preserve similar meanings across languages (or denote identical entities), they may appear in distinct contextual distributions.
Moreover, some homographs correspond to \emph{false friends}, carrying different meanings despite their identical surface forms.
For example, the form \textit{bank} in English can refer to a financial institution or the side of a river, as in \say{They sat by the river bank to rest.} In German, however, \textit{bank} commonly means a bench, as in \say{Ich bin auf der Bank mit dem Kopf gestoßen.} (I hit my head on the bench).
False friends introduce additional complexity to multilingual tokenization: how can a single shared token represent language-dependent meanings and usage patterns?

We begin by showing that mainstream tokenizer algorithms, such as Byte-Pair Encoding~\citep[BPE;][]{sennrich-etal-2016-neural} and Unigram Language Modeling~\citep[ULM;][]{kudo-2018-subword}, often treat homographs in a language-agnostic manner during vocabulary construction.
Because the tokenizer is trained on a concatenated multilingual corpus without explicit language conditioning, identical surface forms are typically assigned a single shared subword unit.
As a result, the downstream embedding representation aggregates signals from multiple languages, potentially reinforcing the imbalance we described.
In contrast, the SaGe tokenizer~\citep{yehezkel-pinter-2023-incorporating}, which incorporates contextual information during vocabulary construction, creates vocabularies that segment shared surface forms differently across languages more often.

To improve homograph handling even more, we introduce \emph{language cues} into the tokenizer training process.
These cues are language-specific Unicode characters injected into the training corpus.
For each shared homograph, the first character is mapped to a language-unique cue, allowing the tokenizer to observe minimal but systematic language distinctions during vocabulary learning.
The multilingual tokenizer is thus encouraged to distinguish homographs across languages, potentially allocating distinct subword units per language and enabling more balanced embeddings.

To examine whether this approach is useful beyond vocabulary construction itself, we evaluate the proposed language cues in a multilingual machine translation setting.
We compare baseline and cue-based English-to-X translation models across several target languages, using both BPE and ULM tokenization and evaluating on general as well as homograph-focused test sets.
While the results do not suggest a uniform improvement in every setting, they indicate that language cues can lead to small and encouraging gains, particularly in some evaluation conditions involving homographs and on more challenging benchmarks.
These findings provide preliminary evidence that introducing lightweight language signals during tokenization may help mitigate problematic shared-token behavior in multilingual downstream tasks.

%% file: fig_cues_intro.tex
\begin{figure}[t]
    \centering
    \includegraphics[width=0.4\textwidth]{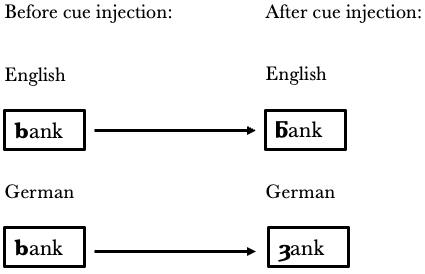}
    \caption{Illustration of language cues applied to a shared homograph before tokenizer training.}
    \label{fig:language_cues_intro}
\end{figure}

%% file: 02_analysis.tex
Over the years, a range of approaches has been proposed to improve multilingual language models.
Studies have looked into specialized vocabulary allocation schemes, composition of multilingual tokenizers from monolingual ones, increasing multilingual vocabularies, alternative encoding methods, vocabulary extension, and data-balancing or adaptation strategies for low-resource languages.
Together, these methods address important challenges such as vocabulary bottlenecks, unfair allocation across languages, and weak performance on underrepresented languages.
At the same time, they mostly treat multilingual tokenization as a problem of global vocabulary design, corpus balance, or model adaptation.

One common assumption in multilingual language modeling is that some degree of vocabulary overlap across languages is useful.
\citet{pires-etal-2019-multilingual} show that multilingual BERT \citep{devlin-etal-2019-bert} can transfer knowledge between languages even without explicit cross-lingual training, which suggests that shared structure can be useful in multilingual models.
\citet{limisiewicz-etal-2023-tokenization} show that vocabulary overlap can be beneficial in some settings, particularly for sentence-level cross-lingual tasks, even if it is not equally helpful for all tasks.
\citet{kallini2025falsefriendsfoesinvestigating} refine this picture by asking what kind of overlap is actually helpful.
They show that overlap is usually better than having completely separate vocabularies, but mostly for words with similar meanings.
Overlap between words with different meanings is much less useful , highlighting the concern for the cases of homographs and false friends.

\input{tab_bpe_v_bpes_ff}

Recently, \citet{kitamura-etal-2025-doppelganger} and \citet{tanwar2026multilingualllmsstrugglelink} showed that models often struggle to disambiguate cross-lingual homographs and may rely too heavily on orthographic similarity rather than semantic interpretation, adding to the motivation of the present work.
We do not treat token sharing only as a question of vocabulary size, clustering, or vocabulary allocation.
Instead, we focus on cases where identical surface forms may be tokenized in the same way, even when their meanings or usage differ across languages.
Our goal is not to eliminate overlap altogether, but to provide the tokenizer with a signal so that potentially misleading shared forms can be distinguished earlier in the pipeline.
In that sense, the proposed language cues operate at a different level from approaches such as \citet{lample2019crosslinguallanguagemodelpretraining}'s, where language embeddings are introduced on the model side during pretraining.
Here, the language signal is introduced directly into the input data, at the point where shared subword units are first created.

\subsection{Analysis of Homograph Tokenization}

To further motivate the proposed approach, we first examine how cross-lingual homographs are tokenized by existing subword tokenizers in monolingual and bilingual settings.
If multilingual tokenizers tend to segment homographs in the same way as their monolingual counterparts, even when meanings differ across languages, this would suggest that the same written form is being segmented similarly regardless of language.
Such a pattern would support the need for an explicit language signal during tokenizer training.

We use a false friends dataset~\citep{aari1995_false_friends_en_de, wiktionary_false_friends} to focus on the clearest case in which identical surface forms should not necessarily be tokenized in the same way across languages.
\autoref{tab:bpe_vs_bpesage_compact_ff} compares monolingual and multilingual tokenizers on this subset under two settings: BPE,\footnote{SentencePiece version~\citep{kudo2018sentencepiecesimplelanguageindependent}.} and SaGe with a BPE-initialized vocabulary.
All tokenizers are trained on monolingual and concatenated multilingual training corpora~\citep{goldhahn-etal-2012-building,leipzig_all_corpora_used}.
The token sequence triplets resulting from each setting (English, L2 $\in\{$German, French$\}$, bilingual) are grouped into several cases: \texttt{same\_splits} denotes cases where the English monolingual, L2 monolingual, and multilingual tokenizers all produce the same segmentation; \texttt{en\_t=mult\_t} and \texttt{L2\_t=mult\_t} denote cases where the multilingual tokenizer matches only one of the monolingual tokenizers; \texttt{en\_t=L2\_t} denotes cases where the two monolingual tokenizers agree with each other but differ from the multilingual tokenizer; and \texttt{different\_splits} denotes cases where all three tokenizations differ.

The comparison shows a clear contrast between tokenizer families.
Standard BPE often produces the same segmentation in monolingual and multilingual settings, while SaGe diverges more strongly.
To determine whether this pattern extends beyond the false-friend subset, we next examine the broader set of homographs.
We build a homograph dataset using a dictionary repository~\citep{altun_all_words_languages}, adding four more languages to our set, and use Earth mover's distance (EMD) to compare the tokenization distributions produced by the BPE and SaGe tokenizer families.
Unlike exact category counts, EMD provides a single measure of how different the overall tokenization patterns are.
Our EMD function is detailed in \autoref{app:emd}.

\input{tab_bpe_v_bpes}

\autoref{tab:bpe_vs_bpesage_compact} shows that the broader homograph set follows the same general pattern observed for false friends.
Across all language pairs, BPE produces many more cases in the \texttt{same\_splits} category, indicating that monolingual and multilingual tokenizers often preserve the same segmentation.
In contrast, the SaGe variant yields substantially more cases in the \texttt{different\_splits} category, suggesting greater divergence between monolingual and multilingual tokenization.
UnigramLM (\autoref{tab:bpe_vs_bpesage_compact}, right) follows a similar pattern to BPE, with monolingual and multilingual tokenizers often producing the same segmentation for homographs, and SaGe reversing the trend.
Taken together, these results suggest that standard BPE and ULM tend to treat cross-lingual homographs in a more language-agnostic way.

%% file: tab_bpe_v_bpes_ff.tex
\begin{table*}[t]
\centering
\small
\begin{tabular}{llccccc}
\toprule
Language Pair & Tokenizer & EN=Mult & L2=Mult & EN=L2 & Same & Diff \\
\midrule
\multirow{2}{30pt}{EN-DE} & BPE       & 21 & 16 & 0 & 33 & 3 \\
      & BPE\_SaGe & 14 & 9  & 8 & 16 & 26 \\
\midrule
\multirow{2}{30pt}{EN-FR} & BPE       & 6  & 6  & 1 & 22 & 5 \\
      & BPE\_SaGe & 9  & 5  & 2 & 12 & 12 \\
\bottomrule
\end{tabular}
\caption{Distribution of tokenization patterns for false friends under BPE and BPE\_SaGe.}
\label{tab:bpe_vs_bpesage_compact_ff}
\end{table*}

%% file: tab_bpe_v_bpes.tex
\begin{table*}[t]
\centering
\small
\begin{tabular}{lHcccccccccccc}
\toprule
& & \multicolumn{6}{c}{BPE} & \multicolumn{6}{c}{ULM} \\
\cmidrule(r){2-8} \cmidrule(l){9-14}
Language & Tokenizer & EN & L2 & EN & All & All & EMD & EN & L2 & EN & All & All & EMD \\
Pair & & =M & =M & =L2 & Same & Diff & & =M & =M & =L2 & Same & Diff \\
\midrule
EN-DE & BPE       & 1924 & 1314 & 170 & 2105 & 894  & \multirow{2}{*}{0.66} & 2001 & 1346 & 191 & 2008 & 861  & \multirow{2}{*}{0.73} \\
     ~~w/SaGe & BPE\_SaGe & 1141 & 750  & 470 & 510  & 3536 & & 1218 & 560  & 278 &  421 & 3930 &  \\
\midrule
EN-FR & BPE       & 1263 & 1055 & 190 & 1853 & 801  & \multirow{2}{*}{0.52}  & 1217 & 1280 & 128 & 1810 &  727 & \multirow{2}{*}{0.67} \\
      ~~w/SaGe & BPE\_SaGe & 797  & 879  & 300 & 786  & 2400 & & 677  &  609 & 508 &  508 & 2860 &                         \\
\midrule
EN-ES & BPE       & 573  & 761  & 69  & 861  & 420  & \multirow{2}{*}{0.54} &  583 &  777 &  63 & 949  &  312 & \multirow{2}{*}{0.65} \\
      ~~w/SaGe & BPE\_SaGe & 440  & 451  & 173 & 304  & 1316 & &  490 &  383 & 216 &  239 & 1356 &                          \\
\midrule
EN-IT & BPE       & 1328 & 565  & 51  & 630  & 809  & \multirow{2}{*}{0.47} & 839 &  1140 &  46 & 936  & 422  & \multirow{2}{*}{0.56} \\
      ~~w/SaGe & BPE\_SaGe & 537  & 283  & 370 & 205  & 1988 &  &  565 & 514  & 305 &  301 & 1698 &                         \\
\midrule
EN-RO & BPE       & 1327 & 712  & 75  & 1533 & 538  & \multirow{2}{*}{0.70} & 947 &  1098 &  114 & 1543 &  483 & \multirow{2}{*}{0.68} \\
      ~~w/SaGe & BPE\_SaGe & 607  & 410  & 461 & 389  & 2318 &  & 607  & 514  & 425 & 419  & 2220 &                         \\
\midrule
EN-SV & BPE       & 606  & 402  & 52  & 932  & 173  & \multirow{2}{*}{0.67} &  652 &  437 & 54  & 821  & 201  & \multirow{2}{*}{0.65} \\
      ~~w/SaGe & BPE\_SaGe & 422  & 268  & 221 & 286  & 968  &  & 380  & 266  & 222 &  255 &  1042 &                        \\
\bottomrule
\end{tabular}
\caption{Distribution of tokenization patterns for homographs under BPE (left) and ULM (right).}
\label{tab:bpe_vs_bpesage_compact}
\end{table*}

%% file: 03_approach.tex
\input{fig_injection}

We introduce lightweight language-specific cues during tokenizer training, at the stage where subword units are learned from text.
First, we construct language-pair-specific homograph sets for English and each target language.
Using these sets, we modify the training corpus by inserting language-specific Unicode cues into selected words, replacing the first character of each marked homograph with a language-unique cue.
We then train tokenizers on the modified corpus and use them in a downstream machine translation setting, where we compare cue-based systems with standard baselines across several English-to-L2 language pairs.

\subsection{Homograph Set Construction}
To construct the homograph sets used in this work, we extracted candidate words from the all-words-in-all-languages repository~\citep{altun_all_words_languages}.
For each English--L2 language pair, we found the intersection between the English word list and the corresponding target-language word list, yielding a set of shared surface forms.
We then applied two filtering steps:
first, we removed words shorter than three characters, in order to avoid irregular forms such as Roman numerals (e.g., ii).
Then, we filtered out words that appeared less than 5 times in either language-specific training corpus. 
This excluded words that are either very rare overall or strongly imbalanced across the two languages, and are therefore less likely to provide reliable evidence during tokenizer training.
The resulting homograph sets are diverse and include names, entities, false friends, and medical terms, among other forms.

\subsection{Language Cues}

Our method introduces language-specific cues directly into the tokenizer training data. \autoref{app:cues} contains the full mapping from lowercase English letters to the language-specific Unicode cue characters we used and explains our selection.
For each language, we define a unique set of 26 Unicode characters, where each cue is mapped to one of the 26 lowercase English letters.\footnote{In all experiments, we lowercased the data in pre-processing.}
This creates a simple one-to-one correspondence between the first letter of a homograph and a language-specific replacement character.
When a word from the homograph set is encountered in the corpus, its first character is replaced with the cue corresponding to that language and letter.

The rationale behind our cue protocol is to provide a lightweight language signal without forcing the two forms into fully separate representations.
Recoverability of the original text is maintained, as the cue replaces only the first character in a systematic one-to-one manner.
We verified that the selected characters are extremely rare in the corpora we used for all tasks, and never exist in beginnings of words where all other characters are standard.
The cue serves as an additional indicator of the linguistic context in which the homograph appears, while the remaining characters can still be shared across languages when the tokenizer finds this useful.
For example, if the English and German forms of \textit{bank} receive different initial cues, the tokenizer may still learn shared subword units such as \textit{ank} or \textit{nk}.
This allows the method to distinguish language-specific uses of a homograph while preserving the possibility of partial cross-lingual sharing.

%% file: fig_injection.tex
\begin{figure*}[t]
    \centering
    \includegraphics[width=0.9\textwidth]{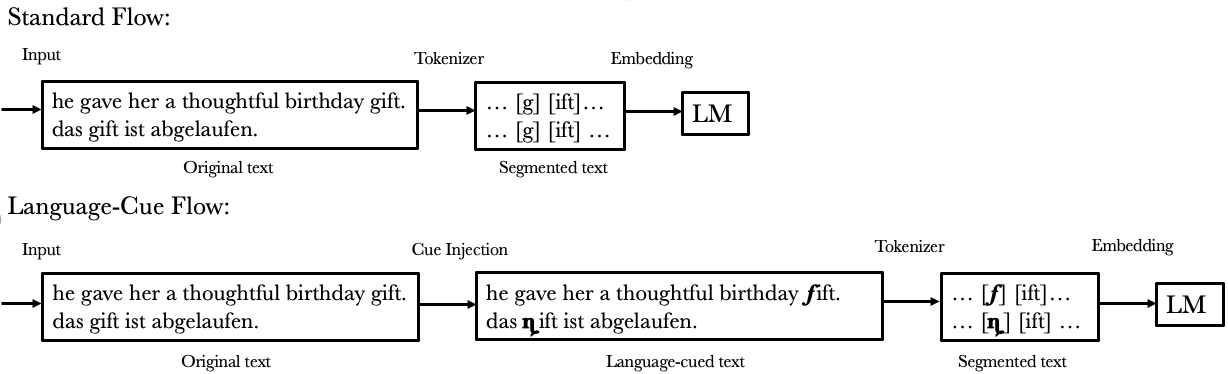}
    \caption{Overview of multilingual language model pipeline: standard flow vs. language-cue injection.}
    \label{fig:flow}
\end{figure*}

%% file: 04_experiments.tex
Before evaluating the proposed method in downstream settings, we examine its static vocabulary features.
We compare multilingual tokenizers trained with language cues against their standard baseline counterparts, with the goal of identifying how cue injection changes basic tokenization behavior.
To characterize these differences, we use several tokenizer-level measures that capture whether cue injection changes the compactness of the learned vocabulary and the granularity of word segmentation. 
\autoref{tab:tokenizer_metrics_de_fr} presents statistics for token length at the vocabulary level and for Rényi efficiency~\citep{zouhar-etal-2023-tokenization} over a held-out development set for English--German and English--French.
Cue-based multilingual tokenizers differ only modestly from their baseline counterparts, but the differences are fairly consistent.\footnote{Similar patterns are observed across the remaining language pairs; the corresponding tables and fertility plots are provided in \autoref{app:langs}.}
In both language pairs, over both tokenizer variants, cued tokens are slightly shorter and efficiency is slightly higher.
The former is expected: as alphabet size increases, more tokens are necessary to encode sequences.
The latter is encouraging: higher Rényi efficiency signifies a more balanced distribution of token types across a corpus.

\input{tab_metrics}

\input{figs_fert}

Figures~\ref{fig:fert_en_de}--\ref{fig:fert_en_fr} compare the \emph{fertility}, presented as tokens per word distributions, between baselines and cue-based multilingual tokenizers.
Cued tokenizers produce fewer single-token words and more words split into three to five subwords, suggesting that cue injection encourages somewhat finer segmentation of the training data.

Although these intrinsic changes are modest, they indicate that cue injection shifts tokenizer behavior in a consistent direction before any downstream training is performed.

\subsection{Language Modeling}

\input{tab_all_ppl}

For each evaluated setup, we train a small GPT language model~\citep{radford2018improving} to evaluate the effect of language cues through perplexity.
Model architecture details are in \autoref{app:lm-deets}.
The models are trained on multilingual text from the OPUS-100~\citep{zhang-etal-2020-improving, tiedemann-2012-parallel} parallel training data, split into 90\% training and 10\% evaluation.
For each tokenizer type, we train one model on the baseline-tokenized data and one model on the cue-tokenized data, and compare their perplexity on the evaluation split.

\autoref{tab:bpe_perplexity} and \autoref{tab:unigram_perplexity} report token-level and word-level perplexity for the baseline and cue-based models. Token-level perplexity is lower for the cue-based models across all language pairs under both BPE and UnigramLM. However, word-level perplexity is generally higher for the cue-based models, with the only exception being EN--IT under UnigramLM, where it improves slightly. Since token-level perplexity depends on segmentation, word-level perplexity provides the more stable comparison in the present setting. Under this normalization, the cue-based models do not show a consistent intrinsic language-modeling advantage. The token-level perplexity results should therefore be interpreted cautiously.

\subsection{Machine Translation}
To further test downstream evaluation, we apply the cue-based tokenization setup to machine translation from English to L2.
For each language pair, we compare two main systems: a baseline model trained on the original parallel data, and a cue-based model trained on a modified version of the same data in which homographs from the constructed set are marked with language-specific cues.
Aside from the cue injection step, the baseline and cue-based systems follow the same training setup.
Models use a shared vocabulary for both languages, and a shared input--output embedding matrix structure, providing a direct assessment of the cue protocol in its intended setting.
We used vocabulary size 8,000.
Further model architecture and implementation details are in \autoref{app:mt-deets}.

\paragraph{Data}
We evaluate the translation models on both standard parallel training data and additional evaluation sets in order to assess the proposed method under different conditions.
As an in-domain benchmark, we use data from OPUS-100~\citep{tiedemann-2012-parallel, zhang-etal-2020-improving}, a large multilingual parallel corpus covering many language pairs.
In addition, we source FLORES+~\citep{nllb-24, flores_plus_hf}, a higher-quality multilingual dataset designed to support more reliable cross-lingual evaluation.
We use it to examine whether the effects of the proposed method remain visible under stricter evaluation conditions.
Finally, from each dataset we derive a \textbf{homograph-focused} subset:
this includes, for each language pair, all and only the sentences that contain at least one homograph from the constructed homograph set on the source side.

\paragraph{Results}

\input{tab_bleu_bpe_all}

\input{tab_bleu_uni_all}

Tables~\ref{tab:bleu_bpe_all} and \ref{tab:bleu_unigram_all} present the BLEU scores for baseline and cued systems across the general and homograph-focused evaluation sets.
Overall, the effect of cue injection is mixed, but several encouraging patterns emerge.
Under BPE, the cued system improves over the baseline in many settings.
These improvements are usually small, but they appear across both the full test sets and the homograph-focused subsets.
At the same time, there are still cases in which the baseline remains stronger, such as EN--DE on OPUS-100 and EN--SV on FLORES+.
A second pattern is that, for BPE, the gains are often more visible on FLORES+ than on OPUS-100.
This is especially noticeable for EN--IT, EN--RO, and EN--DE.
The homograph-focused subsets generally follow the same direction as the full test sets, although they do not consistently show larger gains.
This suggests that the cues may influence translation behavior beyond the selected homograph cases, but not in a uniformly targeted way.

The UnigramLM results are less stable.
Some language pairs show gains, such as EN--DE and EN--FR on OPUS-100 and EN--ES and EN--RO on FLORES+, but in other settings the baseline remains stronger, particularly for EN--SV.
Compared with BPE, UnigramLM shows a less consistent response to cue injection across languages and evaluation sets, possibly because its probabilistic vocabulary selection makes it less sensitive to inserted cues.

Taken together, these results suggest that our cueing method is promising, but its effect is clearly tokenizer- and language-dependent.
BPE appears to benefit more consistently from the cues, while UnigramLM produces a more mixed picture.
Although the gains are modest, their repeated appearance across several settings supports the idea that lightweight language signals at the tokenizer level can have a meaningful downstream effect.

\paragraph{Statistical Robustness}

For statistical robustness, we apply paired bootstrap resampling with 1,000 iterations for each evaluation set.
In each case, we compute the bootstrap distribution of the BLEU difference, defined as cue BLEU minus baseline BLEU, and report its mean, standard deviation, and 95\% confidence interval.
This allows the observed BLEU differences to be examined beyond a single point estimate and helps assess whether the direction of the effect remains stable under repeated resampling.
The full results are in \autoref{app:bootstrap}.
On BPE, two confidence intervals out of 24 do not cross zero, favoring the cued model: English--Italian on both FLORES+ and its homograph subset (mean diff 0.71 and 0.75, resp.).
In UnigramLM, three confidence intervals favor the base model: English--Romanian on OPUS-100 ($-$0.82), and English--Swedish on FLORES+ and its homograph subset ($-$2.28 and $-$2.17, resp.).
The aggregate means for the bootstrap setup are $+$0.1724 for BPE, and $-$0.2495 for UnigramLM, greatly dominated by the two Swedish FLORES+ datasets (without which the difference drops to 0.0699).

The BPE results seem to suggest that the cue-based system often moves in a favorable direction, echoing the relatively stronger results (compared to UnigramLM) on R{\'e}nyi efficiency and perplexity.
UnigramLM's results do not replicate the overall success cues had on intrinsic and language modeling benchmarks.
While the differences are still small in magnitude, the number of language pairs and setups considered allow us to speculate as to the reason these two algorithms behave differently:
BPE is bottom-up, merging tokens based on their components' co-occurrence statistics, which focuses the cues' role as alphabet items that follow similar patterns as the characters they replaced; tokens with cues will form at a slower rate than tokens with the original characters but at a similar order, which will eventually lend consistency to the resulting vocabulary.
UnigramLM, on the other hand, is a top-down algorithm prioritizing candidate token likelihood, which causes it to be less sensitive to low-frequency tokens like the ones cues appear in, making its behavior on them more erratic and denying them consistency.
The cues' differentiating role may be enough to improve intrinsic statistics and LM prediction, but the semantics-heavy task of translation requires more.

\paragraph{Instance Analysis}

It is difficult to find examples in well-established translation datasets where false friends occur in contexts that can throw English-to-L2 translations off.
However, we found several instances where homograph words were translated into native words in the cued models, where the baseline kept the source form, as borrowed-from-English variants.
One example is the word \emph{tour} in the English sentence \textcolor{blue}{\textit{the rock band was due to tour the united states and canada until september 16.}}
The baseline Spanish model translated this to \textcolor{brown}{\textit{la banda de rock se debió al \textbf{tour} a los estados unidos y a canadá hasta el 16 de septiembre.}}, while the cued model produced \textcolor{purple}{\textit{la banda de rock se debía a la \textbf{gira} por los estados unidos y canadá hasta el 16 de septiembre.}}
In another example, the English \textcolor{blue}{\textit{italy's national football, along with german national football team is the second most successful team in the world and were the fifa world cup champions in 2006.}} was translated by the baseline Italian model to \textcolor{brown}{\textit{il calcio nazionale dell'italia, insieme alla nazionale di calcio tedesca, è il secondo \textbf{team} di successo nel mondo ed è stato il campionato della fifa nel 2006.}}, with the cued model preferring \textcolor{purple}{\textbf{\textit{squadra}}}.

%% file: tab_metrics.tex
\begin{table*}[t]
    \centering
    \small
    \begin{tabular}{lcccc}
        \toprule
        & \multicolumn{2}{c}{BPE} & \multicolumn{2}{c}{UnigramLM} \\
        \cmidrule(lr){2-3} \cmidrule(lr){4-5}
        Tokenizer & Avg Len (Vocab) & Rényi Efficiency & Avg Len (Vocab) & Rényi Efficiency \\
        \midrule
        EN           & 5.9008 & 0.4791 & 6.7975 & 0.4565 \\
        \midrule
        DE           & 5.7986 & 0.5333 & 6.7815 & 0.5036 \\
        EN--DE        & 5.4796 & 0.5424 & 6.5370 & 0.4994 \\
        Cued EN--DE   & 5.3039 & 0.5480 & 6.4212 & 0.5028 \\
        \midrule
        FR           & 5.9271 & 0.4941 & 6.8727 & 0.4699 \\
        EN--FR        & 5.6342 & 0.5347 & 6.6221 & 0.4874 \\
        Cued EN--FR   & 5.4085 & 0.5405 & 6.5432 & 0.4873 \\
        \bottomrule
    \end{tabular}
    \caption{Vocabulary length and Rényi efficiency for English--German and English--French tokenizers.}
    \label{tab:tokenizer_metrics_de_fr}
\end{table*}

%% file: figs_fert.tex
\begin{figure}[t]
    \centering
    \includegraphics[width=0.45\textwidth]{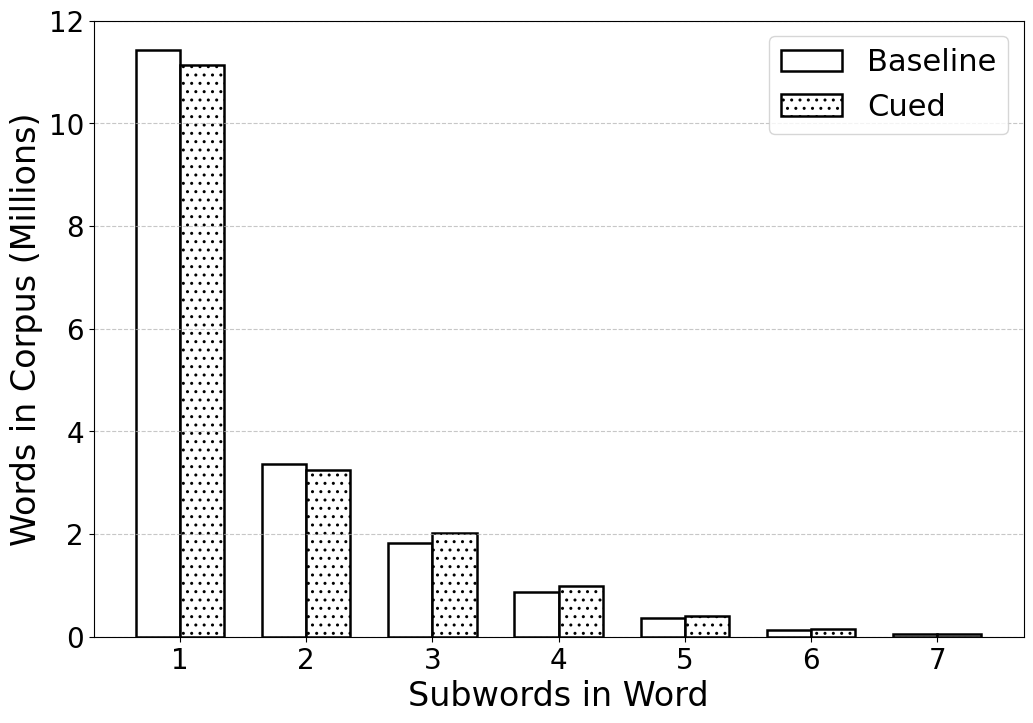}

    \vspace{0.5cm}

    \includegraphics[width=0.45\textwidth]{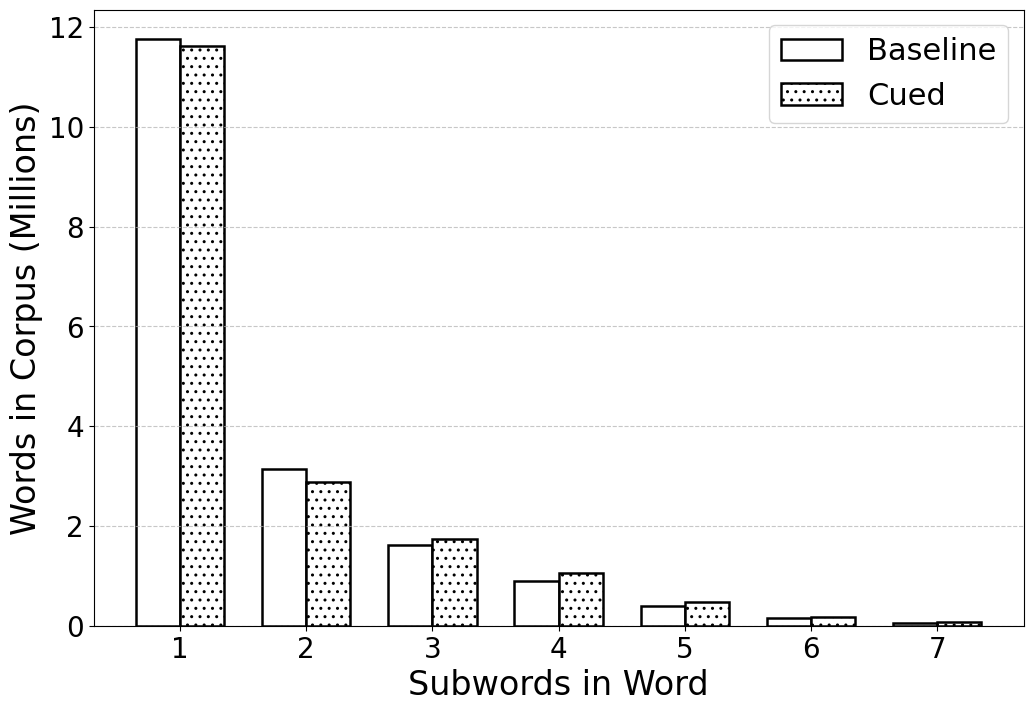}
    \caption{Token fertility comparison between baseline and cue-based multilingual tokenizers for EN--DE, using BPE (top) and UnigramLM (bottom).}
    \label{fig:fert_en_de}
\end{figure}

\begin{figure}[t]
    \centering
    \includegraphics[width=0.45\textwidth]{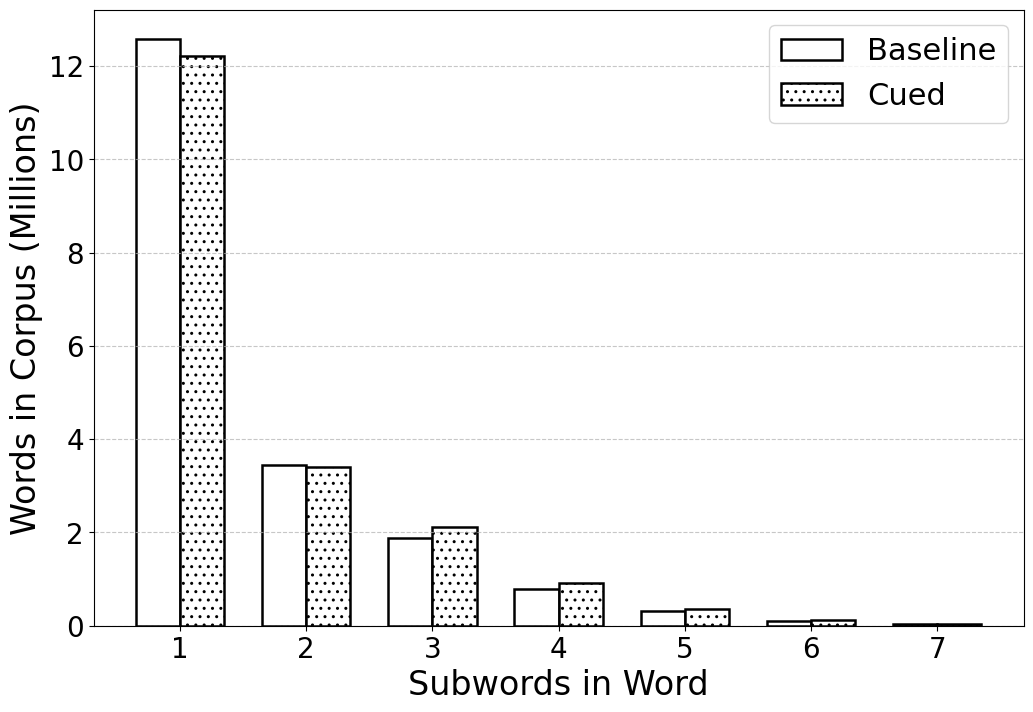}

    \vspace{0.5cm}

    \includegraphics[width=0.45\textwidth]{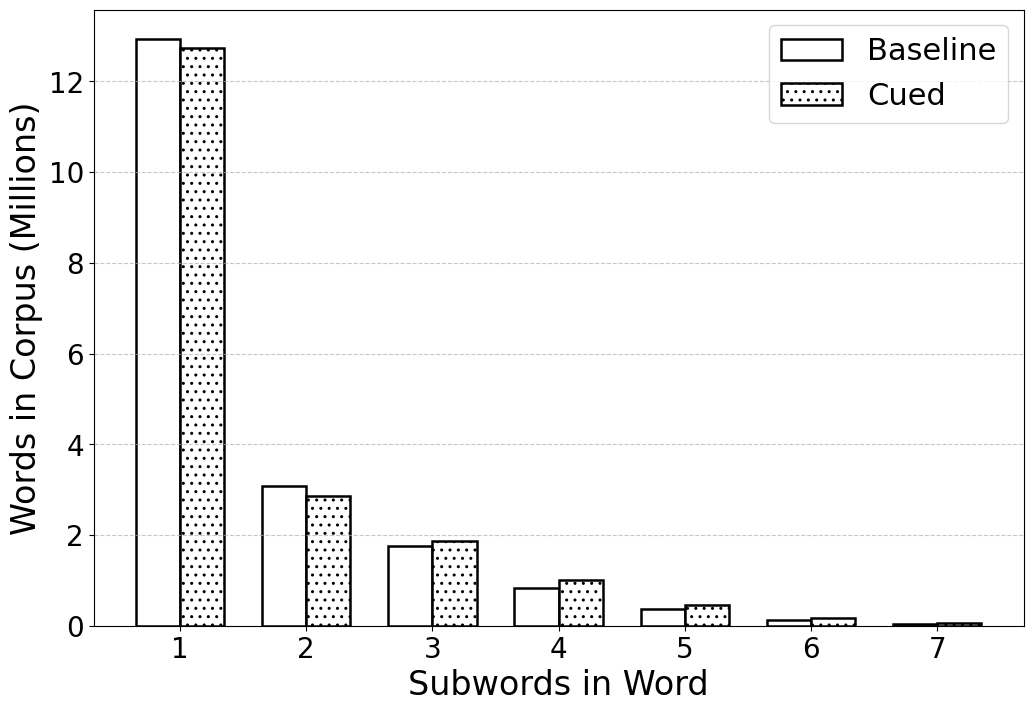}
    \caption{Token fertility comparison between baseline and cue-based multilingual tokenizers for EN--FR, using BPE (top) and UnigramLM (bottom).}
    \label{fig:fert_en_fr}
\end{figure}

%% file: tab_all_ppl.tex

\begin{table}[t]
    \centering
    \small
    \begin{tabular}{lcccc}
        \toprule
        & \multicolumn{2}{c}{Token PPL} & \multicolumn{2}{c}{Word PPL} \\
        \cmidrule(lr){2-3} \cmidrule(lr){4-5}
        L2 & Baseline & Cued & Baseline & Cued \\
        \midrule
        DE & 31.1520 & \textbf{27.8722} & \textbf{581.260} & 593.840 \\
        FR & 24.4689 & \textbf{22.4500} & \textbf{191.658} & 200.497 \\
        ES & 23.7754 & \textbf{22.9584} & \textbf{166.750} & 172.643 \\
        IT & 27.2915 & \textbf{26.1409} & \textbf{251.681} & 260.391 \\
        RO & 19.8321 & \textbf{18.7901} & \textbf{192.845} & 200.475 \\
        SV & 24.1290 & \textbf{22.9260} & \textbf{193.275} & 199.517 \\
        \bottomrule
    \end{tabular}
    \caption{BPE token-level and word-level perplexity results for baseline and cued EN--L2 models. Bolded values indicate the lower perplexity for each language pair and metric.}
    \label{tab:bpe_perplexity}
\end{table}

\begin{table}[t]
    \centering
    \small
    \begin{tabular}{lcccc}
        \toprule
        & \multicolumn{2}{c}{Token PPL} & \multicolumn{2}{c}{Word PPL} \\
        \cmidrule(lr){2-3} \cmidrule(lr){4-5}
        L2 & Baseline & Cued & Baseline & Cued \\
        \midrule
        DE & 27.3393 & \textbf{25.4110} & \textbf{478.566} & 522.405 \\
        FR & 23.7659 & \textbf{21.7882} & \textbf{187.079} & 200.465 \\
        ES & 23.0398 & \textbf{22.2288} & \textbf{155.900} & 161.610 \\
        IT & 27.5019 & \textbf{25.6572} & 254.582 & \textbf{252.936} \\
        RO & 18.9927 & \textbf{18.6392} & \textbf{191.166} & 209.748 \\
        SV & 22.2304 & \textbf{20.9472} & \textbf{175.844} & 185.599 \\
        \bottomrule
    \end{tabular}
    \caption{UnigramLM token-level and word-level perplexity results for baseline and cued EN--L2 models. Bolded values indicate the lower perplexity for each language pair and metric.}
    \label{tab:unigram_perplexity}
\end{table}

%% file: tab_bleu_bpe_all.tex
\begin{table}[t]
    \centering
    \small
    \begin{tabular}{llcc}
        \toprule
        Language & Dataset & Baseline & Cued \\
        Pair \\
        \midrule
        EN-DE & OPUS-100 & \textbf{30.6267} & 30.3367 \\
              & OPUS-100 HG & \textbf{30.1053} & 30.0800 \\
              & FLORES+ & 24.7286 & \textbf{25.0721} \\
              & FLORES+ HG & 24.6836 & \textbf{25.0575} \\
        \midrule
        EN-FR & OPUS-100 & 34.5221 & \textbf{34.8798} \\
              & OPUS-100 HG & 34.7145 & \textbf{35.1657} \\
              & FLORES+ & 38.2848 & \textbf{38.3229} \\
              & FLORES+ HG & 38.2511 & \textbf{38.3176} \\
        \midrule
        EN-ES & OPUS-100 & 37.2895 & \textbf{37.4394} \\
              & OPUS-100 HG & \textbf{37.5043} & 37.4970 \\
              & FLORES+ & 21.8633 & \textbf{21.9806} \\
              & FLORES+ HG & 21.9190 & \textbf{21.9801} \\
        \midrule
        EN-IT & OPUS-100 & 30.4521 & \textbf{30.7234} \\
              & OPUS-100 HG & 30.0807 & \textbf{30.2607} \\
              & FLORES+ & 21.6965 & \textbf{22.4155} \\
              & FLORES+ HG & 21.7195 & \textbf{22.4648} \\
        \midrule
        EN-RO & OPUS-100 & 31.6562 & \textbf{31.7001} \\
              & OPUS-100 HG & \textbf{31.9464} & 31.8473 \\
              & FLORES+ & 21.6225 & \textbf{22.0922} \\
              & FLORES+ HG & 21.6266 & \textbf{22.1890} \\
        \midrule
        EN-SV & OPUS-100 & 31.7053 & \textbf{31.7964} \\
              & OPUS-100 HG & 31.2160 & \textbf{31.3022} \\
              & FLORES+ & \textbf{29.4988} & 29.4864 \\
              & FLORES+ HG & \textbf{30.1117} & 30.0302 \\
        \bottomrule
    \end{tabular}
    \caption{BLEU scores for BPE across the general and homograph-focused test sets (HG).}
    \label{tab:bleu_bpe_all}
\end{table}

%% file: tab_bleu_uni_all.tex
\begin{table}[t]
    \centering
    \small
    \begin{tabular}{llrr}
        \toprule
        Language & Dataset & Baseline & Cued \\
        Pair \\
        \midrule
        EN-DE & OPUS-100 & 31.1366 & \textbf{31.3120} \\
              & OPUS-100 HG & 30.5059 & \textbf{30.8958} \\
              & FLORES+ & \textbf{25.0589} & 24.9009 \\
              & FLORES+ HG & \textbf{25.0610} & 24.8584 \\
        \midrule
        EN-FR & OPUS-100 & 34.6798 & \textbf{35.1994} \\
              & OPUS-100 HG & 34.8713 & \textbf{35.5586} \\
              & FLORES+ & \textbf{38.2869} & 38.0379 \\
              & FLORES+ HG & \textbf{38.3193} & 37.9941 \\
        \midrule
        EN-ES & OPUS-100 & \textbf{37.7815} & 37.4928 \\
              & OPUS-100 HG & \textbf{38.2776} & 37.7797 \\
              & FLORES+ & 21.5442 & \textbf{22.0070} \\
              & FLORES+ HG & 21.6744 & \textbf{22.0411} \\
        \midrule
        EN-IT & OPUS-100 & \textbf{30.6895} & 30.6468 \\
              & OPUS-100 HG & 30.2316 & \textbf{30.3133} \\
              & FLORES+ & \textbf{21.9172} & 21.8406 \\
              & FLORES+ HG & \textbf{21.9610} & 21.8537 \\
        \midrule
        EN-RO & OPUS-100 & \textbf{32.2187} & 31.5070 \\
              & OPUS-100 HG & \textbf{32.3610} & 31.6651 \\
              & FLORES+ & 21.9333 & \textbf{22.1741} \\
              & FLORES+ HG & 21.9793 & \textbf{22.1883} \\
        \midrule
        EN-SV & OPUS-100 & \textbf{31.5568} & 31.1292 \\
              & OPUS-100 HG & \textbf{30.8937} & 30.5175 \\
              & FLORES+ & \textbf{30.1542} & 27.8777 \\
              & FLORES+ HG & \textbf{30.5432} & 28.3570 \\
        \bottomrule
    \end{tabular}
    \caption{BLEU scores for UnigramLM across the general and homograph-focused test sets (HG).}
    \label{tab:bleu_unigram_all}
\end{table}

%% file: 06_conclusion.tex
Multilingual tokenization relies on shared vocabularies in order to represent many languages within a limited number of subword units.
While this sharing is often beneficial, it can also create cases in which identical surface forms are treated too uniformly across languages, even when their meanings or usage differ.
We explored this problem through cross-lingual homographs and proposed a simple way to introduce language information earlier in the tokenization process.

The results presented in this work are modest, but generally encouraging.
The proposed language cues do not lead to consistent gains across all settings, and their effect depends on the tokenizer, the language pair, and the evaluation set.
At the same time, the repeated appearance of small improvements suggests that this type of intervention may be capturing a limitation of standard multilingual tokenization.
This leaves several promising directions for future work, such as exploring whether language cues can be developed into a more general tokenization method rather than remaining a targeted intervention, or extending them to multi-script settings.


%% file: 99_appendix.tex
\section{Cue Replacement Mappings}
\label{app:cues}

\input{tab_cues}
\autoref{tab:cue_table} presents the full mapping between English ASCII characters and those for other languages.

The language cues were selected from the Extended Latin Unicode block.
This choice was made for practical reasons related to SentencePiece training.
First, these characters remain compatible with SentencePiece normalization.
Second, they make it possible to keep the split\_by\_unicode\_script argument set to true, which is preferred in order to avoid mixing characters from unrelated scripts within the same token.
At the same time, because the cues belong to a Latin-based block, they can still participate in merge operations during tokenization training.

\section{Earth Mover's Distance}
\label{app:emd}

Earth mover's distance (EMD) is used for comparing two distributions where the \say{distances} between the points on the supports are meaningful.
For geographic histograms, this is the most natural: euclidean distance dictates the amount of effort needed to \say{move earth} from one location to another; EMD then minimizes the total effort needed given the complete distributions.
For less-trivial cases like ours, custom distance functions can be defined.
The function we implement for $d(x,y)$ is given in \autoref{tab:emd}, intended to distance bins that have different counts of identical tokenizations and bring closer the bins where the multilingual vocabulary behaves regularly.

\begin{table}[t]
    \centering
    \small
    \begin{tabular}{lrrrrrr}
    \toprule
    $x$ / $y$ & EN=M & L2=M & EN=L2 & Same & Diff \\
    \midrule
        EN=Mult & 0.0 \\
        L2=Mult & 0.5 & 0.0 \\
        EN=L2 & 0.7 & 0.7 & 0.0 \\
        Same & 1.0 & 1.0 & 1.0 & 0.0 \\
        Diff & 1.0 & 1.0 & 1.0 & 2.0 & 0.0 \\
        \bottomrule
    \end{tabular}
    \caption{Earth Mover's Distance between tokenizer vocabulary consistency bins.}
    \label{tab:emd}
\end{table}


\section{Intrinsic Results on All Language Pairs}
\label{app:langs}
\begin{figure}[t]
    \centering
    \includegraphics[width=0.45\textwidth]{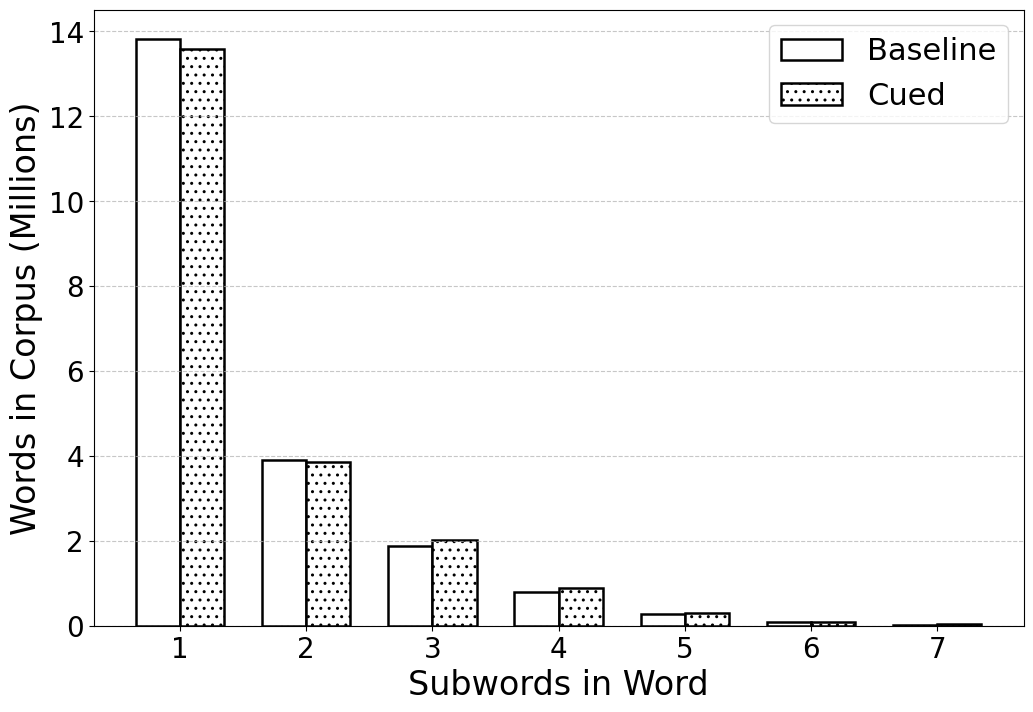}

    \vspace{0.5cm}

    \includegraphics[width=0.45\textwidth]{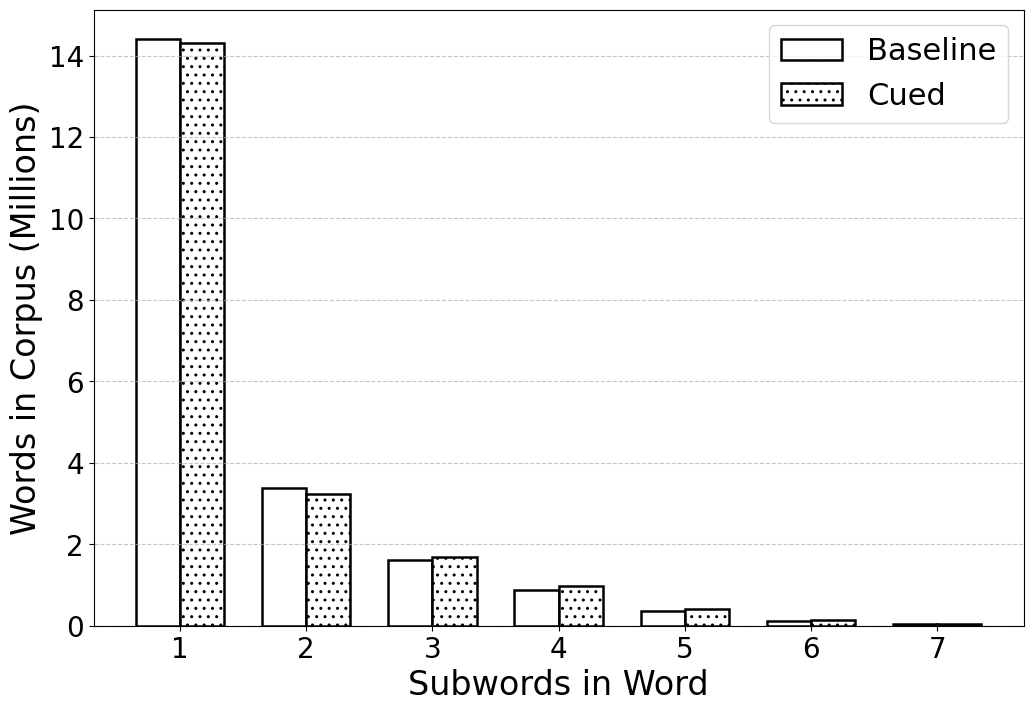}
    \caption{Token fertility comparison between baseline and cue-based multilingual tokenizers for EN--ES, using BPE (top) and UnigramLM (bottom).}
    \label{fig:fert_en_es}
\end{figure}

\begin{figure}[t]
    \centering
    \includegraphics[width=0.45\textwidth]{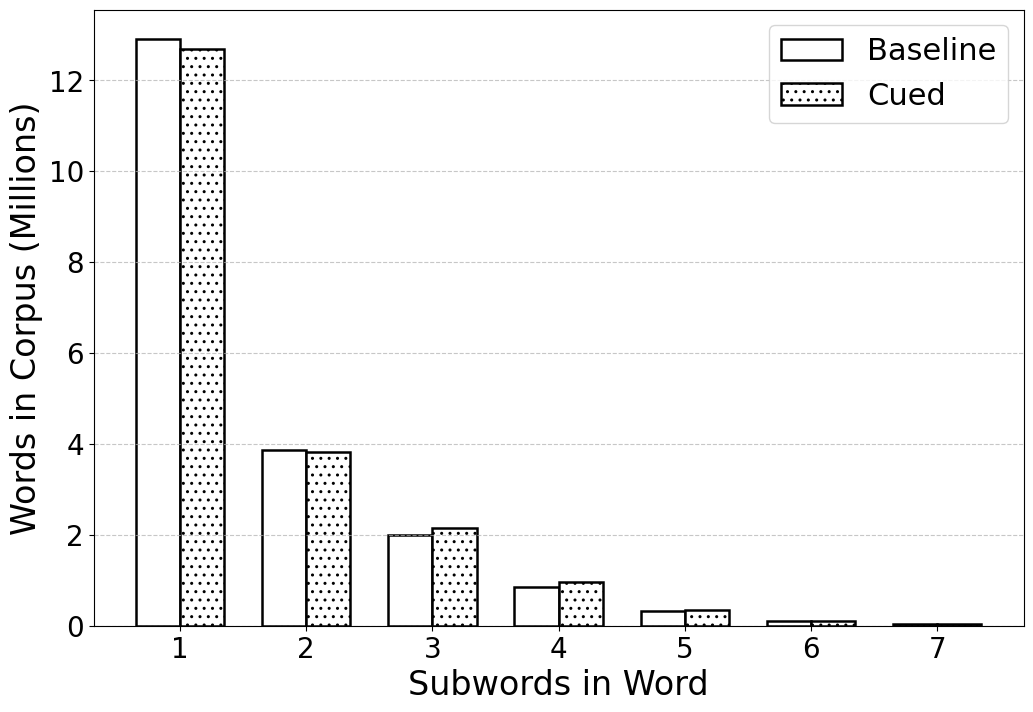}

    \vspace{0.5cm}

    \includegraphics[width=0.45\textwidth]{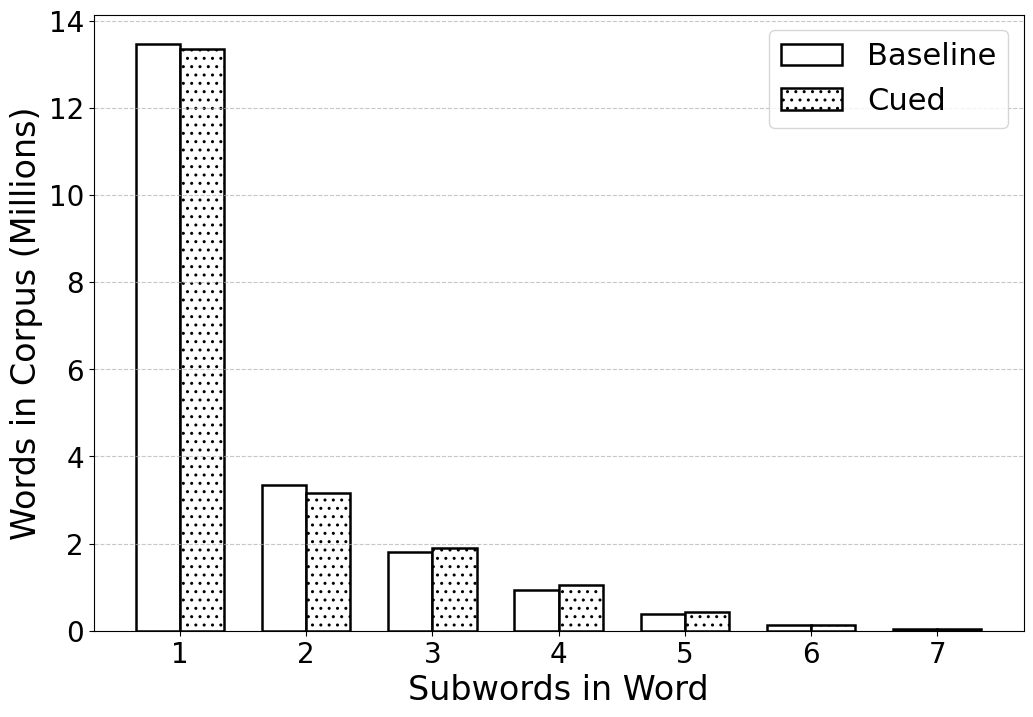}
    \caption{Token fertility comparison between baseline and cue-based multilingual tokenizers for EN--IT, using BPE (top) and UnigramLM (bottom).}
    \label{fig:fert_en_it}
\end{figure}

\begin{figure}[t]
    \centering
    \includegraphics[width=0.45\textwidth]{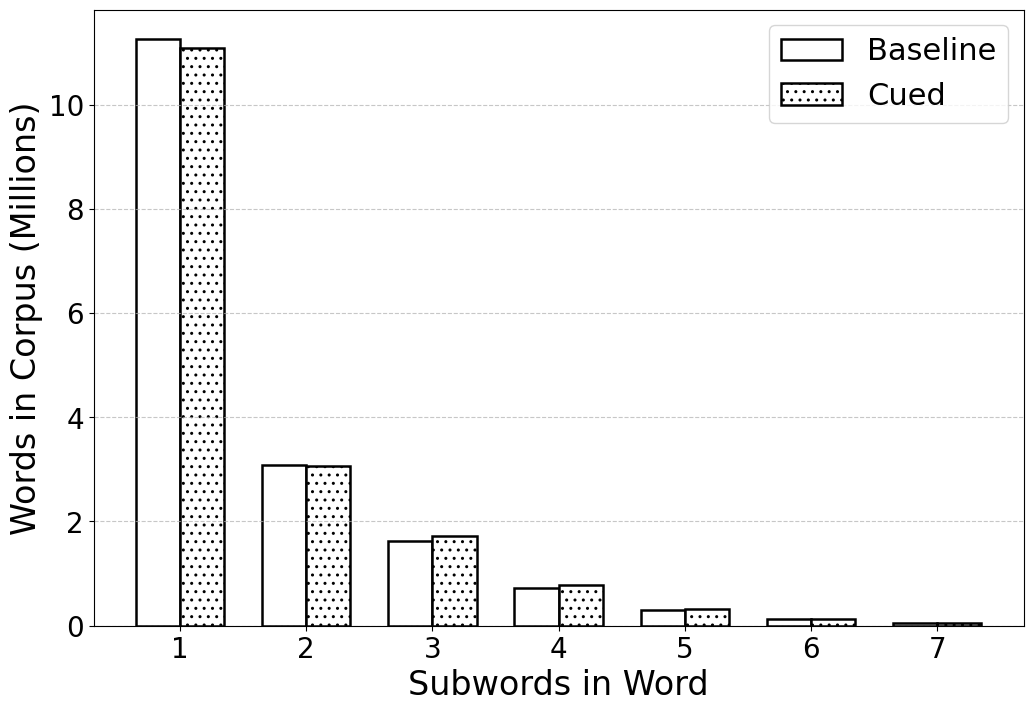}

    \vspace{0.5cm}

    \includegraphics[width=0.45\textwidth]{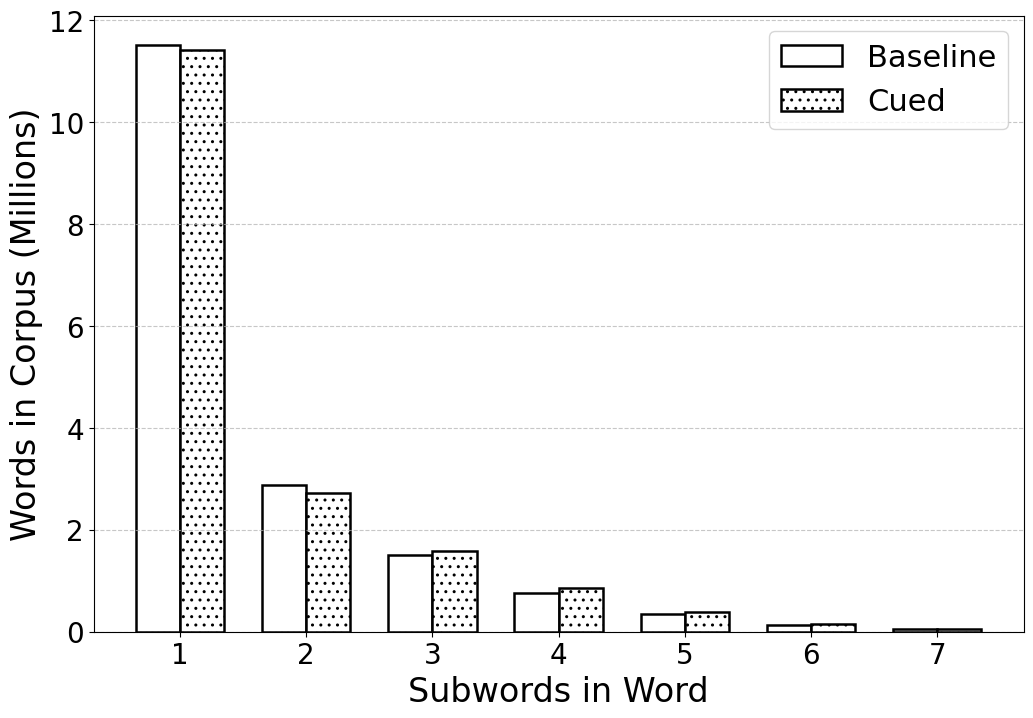}
    \caption{Token fertility comparison between baseline and cue-based multilingual tokenizers for EN--SE, using BPE (top) and UnigramLM (bottom).}
    \label{fig:fert_en_se}
\end{figure}

\begin{figure}[t]
    \centering
    \includegraphics[width=0.45\textwidth]{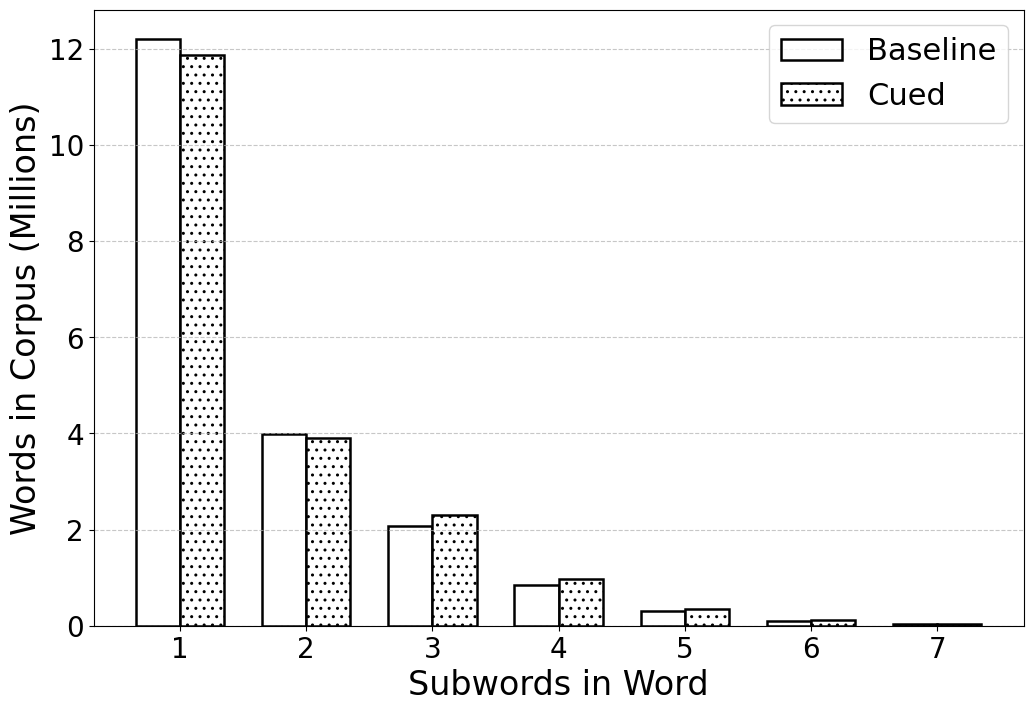}

    \vspace{0.5cm}

    \includegraphics[width=0.45\textwidth]{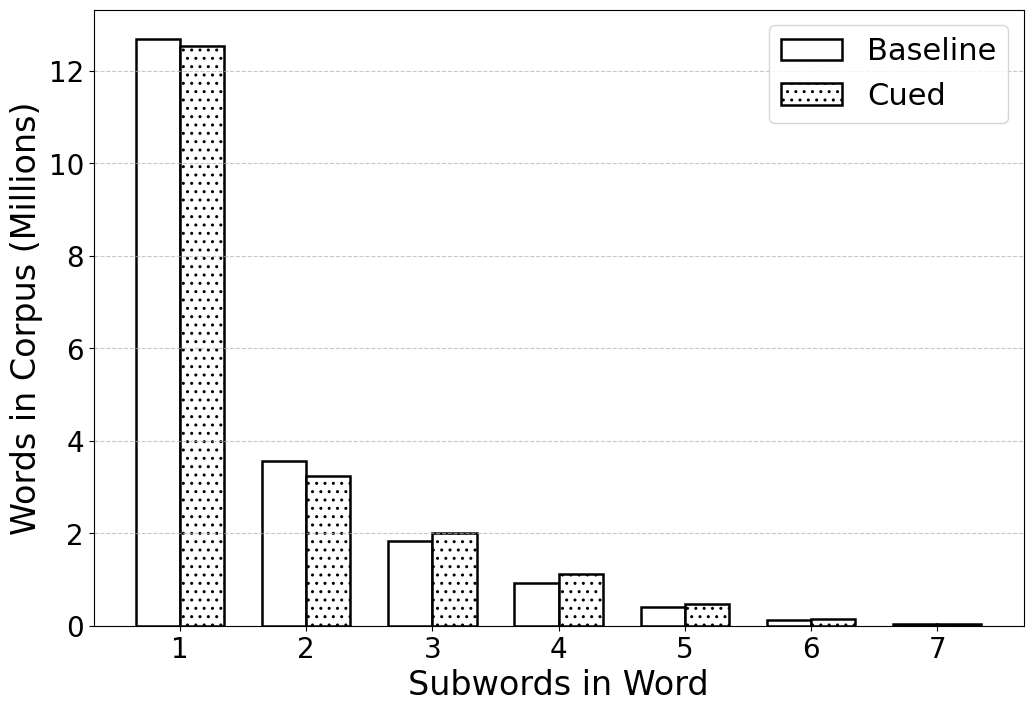}
    \caption{Token fertility comparison between baseline and cue-based multilingual tokenizers for EN--RO, using BPE (top) and UnigramLM (bottom).}
    \label{fig:fert_en_ro}
\end{figure}

Figures~\ref{fig:fert_en_es}, \ref{fig:fert_en_it}, \ref{fig:fert_en_se}, and \ref{fig:fert_en_ro} present the fertility comparisons for English and Spanish, Italian, Swedish, and Romanian, respectively.

\section{GPT Experimental Setup}
\label{app:lm-deets}

The parameters used for the language model experiments are presented in \autoref{tab:gpt_training_params}.

\begin{table}[t]
    \centering
    \small
    \begin{tabular}{ll}
        \toprule
        Parameter & Value \\
        \midrule
        Block size & 128 \\
        Stride & 64 \\
        Vocabulary size & 8{,}000 \\
        Embedding dimension & 256 \\
        Transformer layers & 4 \\
        Attention heads & 8 \\
        Epochs & 3 \\
        Training batch size & 16 \\
        Evaluation batch size & 16 \\
        Learning rate & $5 \times 10^{-4}$ \\
        Weight decay & 0.01 \\
        \bottomrule
    \end{tabular}
    \caption{GPT training and model configuration parameters.}
    \label{tab:gpt_training_params}
\end{table}

\section{NMT Experimental Setup}
\label{app:mt-deets}

All machine translation experiments were conducted using the fairseq-based~\citep{ott2019fairseq} implementation described by \citet{lee-etal-2025-jamo}. 
The only modification made to fairseq was in the post-processing of the validation and test outputs, so that BLEU would be computed on the correctly reconstructed text. Final BLEU scores were computed with SacreBLEU after detokenization, matching the post-processing pipeline used for the generated outputs.
Apart from this step, the training pipeline followed the standard fairseq workflow, with parameters provided in \autoref{tab:fairseq_model_params}.
The tokenizer used in all experiments was SentencePiece with the standard settings.
The translation architecture used in all experiments was transformer\_iwslt\_de\_en.
As a result, each model used a single tokenizer with a shared vocabulary of size 8,000, trained either with BPE or with UnigramLM.
Training took between 2 and 4 hours for all models on a given setup (baseline / cues $\times$ BPE / Unigram) on an NVidia GeForce RTX 4090 GPU server.

\input{tab_fairseq_params}

\section{Bootstrap MT}
\label{app:bootstrap}

Tables~\ref{tab:bootstrap_bpe} and \ref{tab:bootstrap_unigram} present full results for the bootstrap tests on BPE and UnigramLM, respectively.

\input{tab_bootstrap_bpe}

\input{tab_bootstrap_uni}

%% file: tab_cues.tex
\begin{table}[t]
    \centering
    \includegraphics[width=0.4\textwidth]{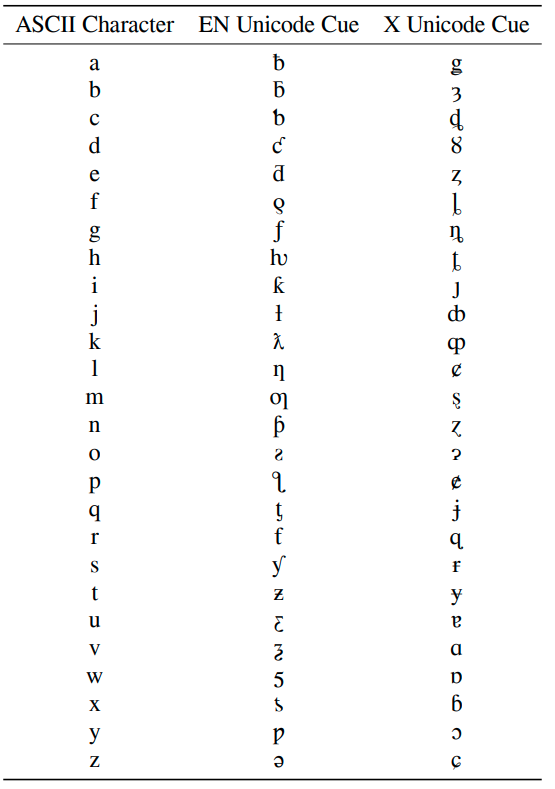}
    \caption{Mapping from English ASCII characters to language-specific Unicode cue replacements. EN denotes the English-side cue, and X denotes the cue used for the non-English language in the pair.}
    \label{tab:cue_table}
\end{table}

%% file: tab_fairseq_params.tex
\begin{table}[t]
    \centering
    \small
    \begin{tabular}{ll}
        \toprule
        Parameter & Value \\
        \midrule
        Architecture & \texttt{transformer\_iwslt\_de\_en} \\
        Embedding dimension & 512 \\
        FFN dimension & 1024 \\
        Attention heads & 4 \\
        Layers  & 6 \\
        Dropout & 0.1 \\
        Vocabulary size & 8{,}000 \\
        Tokenizer software & SentencePiece \\
        Tokenizer & BPE / UnigramLM \\
        Embedding sharing & All embedding matrices shared \\
        Total parameter count & 35.6 million \\
        \bottomrule
    \end{tabular}
    \caption{Fairseq model hyperparameters used in the experiments (Encoder and Decoder).}
    \label{tab:fairseq_model_params}
\end{table}

%% file: tab_bootstrap_bpe.tex
\begin{table*}[t]
    \centering
    \small
    \setlength{\tabcolsep}{4pt}
    \begin{tabular}{llccccc}
        \toprule
        Language & Dataset & Base Mean & Cued Mean & CI & CI Mean & CI STD \\
        Pair \\
        \midrule
        EN-DE & OPUS-100 &  \textbf{30.5622} & 30.2333 & [-1.7228, 1.0483] & -0.3288 & 0.7068   \\
              & OPUS-100 HG & \textbf{30.0941} & 30.01661 & [-1.5112, 1.4028] & -0.0775 & 0.7465 \\
              & FLORES+ & 24.7009 & \textbf{25.0357} & [-0.5123, 1.2048] & 0.3348 & 0.4339 \\
              & FLORES+ HG & 24.6783 & \textbf{25.0447} & [-0.4346, 1.2098] & 0.3663 & 0.4223 \\
        \midrule
        EN-FR & OPUS-100 & 34.477 & \textbf{34.6005} & [-1.0017, 0.9755] & 0.1235 & 0.4991 \\
              & OPUS-100 HG & 34.7592 & \textbf{35.1839} & [-0.2267, 1.172] & 0.4246 & 0.3777 \\
              & FLORES+ & 38.2764 & \textbf{38.317} & [-0.8081, 0.8337] & 0.0406 & 0.4068 \\
              & FLORES+ HG & 38.2576 & \textbf{38.3066} & [-0.7785, 0.8528] & 0.0489 & 0.4138 \\
        \midrule
        EN-ES & OPUS-100 & 37.2876 & \textbf{37.4389} & [-0.4958, 0.8399] & 0.1513 & 0.3419 \\
              & OPUS-100 HG & \textbf{37.5327} & 37.5313 & [-0.7809, 0.8577] & -0.0015 & 0.4277 \\
              & FLORES+ & 21.8463 & \textbf{21.9726} & [-0.4088, 0.6504] & 0.1263 & 0.2687 \\
              & FLORES+ HG & 21.9148 & \textbf{21.9713} & [-0.5555, 0.669] & 0.0565 & 0.3111 \\
        \midrule
        EN-IT & OPUS-100 & 30.4669 & \textbf{30.7343} & [-0.3476, 0.9778] & 0.2673 & 0.3481 \\
              & OPUS-100 HG & 30.0792 & \textbf{30.2757} & [-0.5437, 1.0203] & 0.1965 & 0.3891 \\
              & FLORES+ & 21.5773 & \textbf{22.388} & [0.1406, 1.2879] & 0.7107 & 0.3013 \\
              & FLORES+ HG & 21.7211 & \textbf{22.4678} & [0.1791, 1.3498] & 0.7467 & 0.309 \\
        \midrule
        EN-RO & OPUS-100 & 31.6467 & \textbf{31.6821} & [-0.7409, 0.7931] & 0.0353 & 0.3912 \\
              & OPUS-100 HG & \textbf{31.8597} & 31.7761 & [-1.0102, 0.819] & -0.0836 & 0.47 \\
              & FLORES+ & 21.6031 & \textbf{22.0766} & [-0.2505, 1.1854] & 0.4735 & 0.3727 \\
              & FLORES+ HG & 21.6132 & \textbf{22.1975} & [-0.1953, 1.3399] & 0.5843 & 0.3842 \\
        \midrule
        EN-SV & OPUS-100 & \textbf{31.6725} & 31.6634 & [-0.64, 0.6581] & -0.009 & 0.3471 \\
              & OPUS-100 HG & 31.2091 & \textbf{31.2311} & [-0.7806, 0.8409] & 0.022 & 0.3947 \\
              & FLORES+ & 29.4778 & \textbf{29.4790} & [-0.7923, 0.74077] & 0.0011 & 0.3991 \\
              & FLORES+ HG & 30.0811 & \textbf{30.86} & [-0.8853, 0.7854] & -0.0724 & 0.4326 \\
        \bottomrule
    \end{tabular}
    \caption{Bootstrap BLEU results for BPE across all language pairs and evaluation sets. Bold values mark the better BLEU mean between the baseline and cued systems.}
    \label{tab:bootstrap_bpe}
\end{table*}

%% file: tab_bootstrap_uni.tex
\begin{table*}[t]
    \centering
    \small
    \setlength{\tabcolsep}{4pt}
    \begin{tabular}{llccccc}
        \toprule
        Language Pair & Dataset & Base Mean & Cued Mean & CI & CI Mean & CI STD \\
        \midrule
        EN-DE & OPUS-100 & 31.0801 & \textbf{31.0872} & [-1.0589, 0.8544] & 0.007 & 0.4961 \\
              & OPUS-100 HG & 30.4799 & \textbf{30.788} & [-0.5922, 1.1497] & 0.3081 & 0.4499 \\
              & FLORES+ & \textbf{25.0344} & 24.8776 & [-1.0065, 0.6529] & -0.1568 & 0.4222 \\
              & FLORES+ HG & \textbf{25.0487} & 24.8476 & [-1.0939, 0.7092] & -0.201 & 0.4582 \\
        \midrule
        EN-FR & OPUS-100 & 34.6611 & \textbf{35.0633} & [-0.583, 1.6067] & 0.4022 & 0.5898 \\
              & OPUS-100 HG & 34.9245 & \textbf{35.5875} & [-0.3081, 2.0328] & 0.6629 & 0.6242 \\
              & FLORES+ & \textbf{38.2886} & 38.0368 & [-1.0739, 0.5733] & -0.2517 & 0.4273 \\
              & FLORES+ HG & \textbf{38.3081} & 37.9923 & [-1.2065, 0.5239] & -0.3158 & 0.4383 \\
        \midrule
        EN-ES & OPUS-100 & \textbf{37.7693} & 37.4949 & [-0.914, 0.3236] & -0.2743 & 0.3091 \\
              & OPUS-100 HG & \textbf{38.3088} & 37.8305 & [-1.2022, 0.2749] & -0.4782 & 0.3825 \\
              & FLORES+ & 21.5271 & \textbf{21.9784} & [-0.0441, 0.9376] & 0.4512 & 0.2658 \\
              & FLORES+ HG & 21.6727 & \textbf{22.0392} & [-0.1863, 0.963] & 0.3664 & 0.2913 \\
        \midrule
        EN-IT & OPUS-100 & \textbf{30.6927} & 30.6533 & [-0.7915, 0.7298] & -0.0394 & 0.3796 \\
              & OPUS-100 HG & 30.258 & \textbf{30.3367} & [-0.7897, 1.0717] & 0.0787 & 0.4733 \\
              & FLORES+ & \textbf{21.9055} & 21.8184 & [-0.6979, 0.5808] & -0.0871 & 0.3244 \\
              & FLORES+ HG & \textbf{21.9552} & 21.8576 & [-0.6841, 0.5635] & -0.0975 & 0.3205 \\
        \midrule
        EN-RO & OPUS-100 & \textbf{32.2069} & 31.3905 & [-1.7053, -0.0142] & -0.8164 & 0.4356 \\
              & OPUS-100 HG & \textbf{32.3069} & 31.5472 & [-1.6286, 0.1514] & -0.7596 & 0.462 \\
              & FLORES+ & 21.9203 & \textbf{22.1662} & [-0.4249, 0.943] & 0.2458 & 0.3466 \\
              & FLORES+ HG & 21.9923 & \textbf{22.168} & [-0.5265, 0.8436] & 0.1756 & 0.349 \\
        \midrule
        EN-SV & OPUS-100 & \textbf{31.5023} & 31.084 & [-1.1569, 0.2807] & -0.4182 & 0.3724 \\
              & OPUS-100 HG & \textbf{30.8659} & 30.5252 & [-1.1369, 0.4566] & -0.3406 & 0.4566 \\
              & FLORES+ & \textbf{30.1383} & 27.8624 & [-3.1942, -1.4039] & -2.2759 & 0.4591 \\
              & FLORES+ HG & \textbf{30.5154} & 28.343 & [-3.0399, -1.2409] & -2.1723 & 0.4659 \\
        \bottomrule
    \end{tabular}
    \caption{Bootstrap BLEU results for UnigramLM across all language pairs and evaluation sets. Bold values mark the better BLEU mean between the baseline and cued systems.}
    \label{tab:bootstrap_unigram}
\end{table*}